\documentclass[a4paper, 10pt, conference]{ieeeconf}      
\usepackage{FG2026}
\usepackage{booktabs}
\usepackage{multirow}
\usepackage{amsmath}
\usepackage{graphicx}
\usepackage{algorithm}
\usepackage{algpseudocode}
\usepackage{xcolor}
\usepackage{caption} 
\usepackage{float}   
\usepackage{float}   
\usepackage{hyperref}
\definecolor{qcolor}{HTML}{BFD8F0}
\definecolor{acolor}{HTML}{F5CBAA}
\definecolor{ocolor}{HTML}{D7EEAC}
\FGfinalcopy 

\usepackage{fancyhdr}

\IEEEoverridecommandlockouts                              
\def\FGPaperID{309} 

\title{\LARGE \bf
Elastic Spiking Transformers for Efficient Gesture Understanding
}

\author{\parbox{16cm}{\centering
{\large Alberto Ancilotto$^1$, Gianluca Amprimo$^2$, Stefano Di Carlo$^2$, and Elisabetta Farella$^1$}\\
{\normalsize
$^1$ Fondazione Bruno Kessler \\
$^2$ Politecnico di Torino}}
\thanks{This work was supported by Ministero delle Imprese e del Made in Italy (IPCEI Cloud DM 27 giugno 2022 – IPCEI-CL-0000007) and European Union (Next Generation EU).}
}
\begin{document}

\ifFGfinal
\thispagestyle{empty}
\pagestyle{empty}
\else
\author{Anonymous FG2026 submission\\ Paper ID \FGPaperID \\}
\pagestyle{plain}
\fi
\maketitle

\thispagestyle{fancy}
\renewcommand{\headrulewidth}{0pt}


\begin{abstract}

Spiking Neural Networks (SNNs), and particularly Spiking Transformers, have recently emerged as a promising solution for energy-efficient processing of event-based sensor data in healthcare applications. However, existing architectures are inherently rigid, being trained and deployed as static networks with fixed parameter counts and computational graphs. This rigidity presents a critical bottleneck for deployment on neuromorphic hardware (e.g., Loihi, Spinnaker), where on-chip hardware limitations often force the use of smaller, lower-accuracy models that sacrifice representational capacity and accuracy. To bridge this gap, we introduce the Elastic Spiking Transformer, a novel architecture that embeds runtime adaptability directly into the spiking paradigm. Drawing inspiration from Matryoshka-style representation learning, our framework introduces nested elasticity within the Feature Extractor, Spiking Self-Attention, and Feed-Forward blocks. By training a single ``universal" model with granularity-aware weight sharing, we enable dynamic slicing of network width and attention heads during inference without retraining. This elasticity offers a unique dual advantage in the SNN domain. First, it solves the memory-constrained problem by allowing the model to instantly scale its parameter footprint to fit diverse hardware specifications without retraining. Second, and crucially, we show that elasticity in SNNs not only lowers synaptic weight counts but also directly lowers the total spike firing rate. By dynamically reducing the number of active neurons, our approach achieves a proportional reduction in synaptic operations, unavailable to standard artificial neural networks. We validate our approach on standard benchmarks (CIFAR10/100, CIFAR10-DVS) and the novel EHWGesture clinical gesture understanding dataset. Our results demonstrate that a single Elastic Spiking Transformer spans a wide range of complexity trade-offs, matching or surpassing the accuracy of independently trained baselines while enabling adaptive, real-time gesture recognition on resource-constrained edge devices.

\end{abstract}

\section{INTRODUCTION}

Gesture recognition and understanding are critical tasks for next-generation human–computer interaction, offering a wide range of applications, from clinical rehabilitation monitoring \cite{rehab} and surgical assistance \cite{surgery} to sign language interpretation \cite{rastgoo2021sign} and hands-free device control \cite{yang2023smart}. Deployment on always-on, low-power edge devices allows continuous gesture monitoring, enabling real-time interaction while preserving user privacy and reducing communication latency \cite{dvs128_gesture, gesture-edge}. However, deployment on such devices faces a fundamental challenge, as their strict power budgets and capabilities limit the computational complexity required for accurate gesture understanding.

\begin{figure}[]
  \centering
   \includegraphics[width=\linewidth]{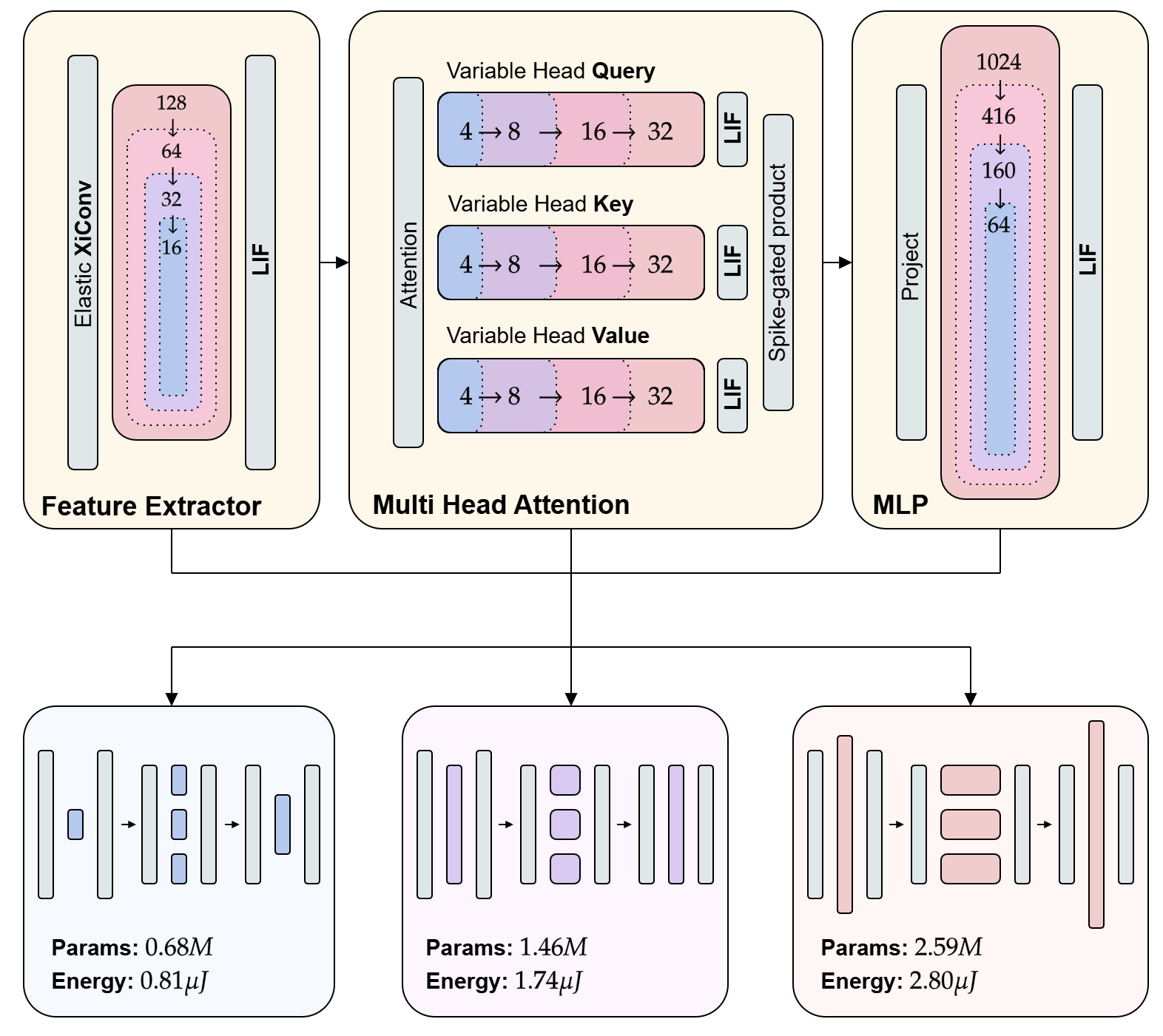}
   \caption{Proposed NESTformer modules.}
   \label{fig:teaser}
\end{figure}

Event cameras offer a promising sensing modality for gesture recognition on such low-resource platforms. Unlike traditional frame-based cameras, these sensors only encode motion, producing a sparse representation that naturally aligns with Spiking Neural Networks (SNNs) based processing. SNNs enable computation only when spikes arrive, reducing costly multiply-accumulate (MAC) operations to simple accumulations. On neuromorphic hardware platforms such as Intel Loihi \cite{intel_loihi} and IBM TrueNorth \cite{ibm_northpole, ibm_truenorth}, this paradigm achieves energy consumption as low as 0.9-26 pJ per synaptic operation, enabling always-on gesture recognition on severely power-constrained devices.

Recent advances in SNNs have brought attention-based architectures to the neuromorphic domain \cite{spikformer, spike-driven-transformer, qkformer}, achieving strong performance on gesture benchmarks. However, despite recent progress, current spiking transformers still have architectural limitations that limit their effective use on edge devices. First, existing models are structurally rigid; these architectures have fixed sizes, meaning moving from one neuromorphic chip to another requires scaling and training a completely new version.
Secondly, these architectures still often rely on CPU processing. For example, many of these architectures rely on matrix multiplication, which is generally incompatible with the spike-driven processing of true neuromorphic chips. As these chips do not handle such global computations natively, the workload must be offloaded to a higher-power CPU, limiting the actual energy savings in real life.

To address these two limitations, we introduce the Neuromorphic Elastic Spiking Transformer (NESTformer) architecture.
Inspired by recent advancements in elastic transformer architectures \cite{matformer}, NESTformer was developed to allow elasticity to spiking transformer architectures.
Our framework is based on three novel architectural blocks:

\begin{enumerate}
    \item \textbf{Elastic Spiking MLP} inspired by Matryoshka-style representation learning \cite{matryoshka_style_representation_learning}, uses a variable hidden dimension, which allows post-train tuning of the model operations and parameter count.
    \item \textbf{Elastic Spiking Attention:} A novel approach to spiking self-attention \cite{spikformer} that allows variable attention head count while also ensuring compatibility when mapping to neuromorphic hardware.
    \item \textbf{Elastic Feature Extraction:} A novel low-resource architecture allows for post-train tuning of the final network energy usage.
\end{enumerate}

This elasticity offers a dual advantage for gesture understanding, unique in the SNN domain. First, it solves the deployment flexibility problem by allowing the model to instantly scale its parameter footprint to fit devices with significantly different hardware resources and power levels, without retraining. Second, we demonstrate that elasticity in SNNs also corresponds to proportional reductions in compute costs: reducing network size not only lowers synaptic weight counts but also directly reduces the total spike firing rate, achieving significantly lower energy consumption than current state-of-the-art approaches at comparable or better accuracies.

The main contributions of this work are:
\begin{itemize} \item \textbf{Elastic spiking transformer architecture:} We propose NESTformer, the first SNN transformer model, to the best of our knowledge, capable of runtime adaptability without retraining. By introducing elasticity across all three building blocks (feature extraction, self-attention, and MLP), we enable a single universal model to dynamically scale its memory footprint without retraining.

\item \textbf{Neuromorphic-native attention mechanism:} We introduce a novel row-wise spiking attention computation scheme that replaces global matrix multiplications with sequential synaptic operations, for better compatibility with neuromorphic-native hardware.

\item \textbf{Energy scaling:} We analyze existing spiking transformers and demonstrate how scaling model size often yields sub-linear energy savings due to high spike generation in the feature extraction stage. In contrast, we show that NESTformer's elasticity directly reduces the total spike firing rate, resulting in a linear reduction in energy.

\item \textbf{State-of-the-Art performance:} We validate our approach on the EHWGesture clinical dataset \cite{ehwgesture} and standard event-based benchmarks (DVS128 \cite{dvs128_gesture}, CIFAR10-DVS \cite{cifar10_dvs}). Our results show that NESTformer matches or outperforms independently trained baselines (75.98\% on EHWGesture, 98.96\% on DVS128) while consuming up to 60\% less energy than current state-of-the-art architectures.
\end{itemize}


\section{RELATED WORKS}

\subsection*{Spiking Neural Networks}

Early SNN research focused on ANN-to-SNN conversion methods \cite{conversion_diehl, conversion_rueckauer} or on directly training Convolutional SNNs using surrogate gradient techniques \cite{surrogate_neftci, surrogate_stbp}.
Recent advances in Spiking Transformers have brought attention-based architectures to the neuromorphic domain, achieving strong performance on gesture benchmarks.  \textit{Spikformer} \cite{spikformer} introduced Spiking Self-Attention (SSA), replacing the softmax operation with spike-based query-key-value interactions.
 \textit{Spikingformer}~\cite{cml} refined residual connections with ``Dual-Spike'' connections to improve gradient flow, enabling deeper networks with improved accuracy. \textit{QKFormer}~\cite{qkformer}, introduced a hierarchical two-stage architecture with Q-K Attention, requiring linear complexity with respect to token length. However, like its predecessors, QKFormer still relies on operations, such as Batched Matrix Multiply (BMM), that are not natively supported on current neuromorphic hardware and instead run on the CPU. \textit{Spike-driven Transformer}~\cite{spike-driven-transformer} instead focused on strict adherence to the spike-driven philosophy, introducing Spike-Driven Self-Attention (SDSA) that eliminates all multiplications, relying solely on mask-and-add operations. While this approach significantly reduces energy consumption, it comes at the cost of reduced representational capacity and lower accuracy compared to BMM-based spiking transformers. 
 Overall, existing spiking transformers face a trade-off between architectural expressiveness and true neuromorphic compatibility, and they are typically designed as fixed-size models without runtime adaptability.

\subsection*{Elastic and Slimmable Networks}

Classical scalable backbones usually require retraining once the architecture has been correctly sized for a target computational budget\cite{mobilenet, phinet, xinet}.
Elastic or slimmable networks allow a single trained model to operate at multiple computational budgets by dynamically adjusting width, depth, or resolution from a given \textit{granularity} setting. \textit{Slimmable Networks}~\cite{slimmable} introduced width-adjustable CNNs that can run at different channel multipliers (0.25$\times$--1.0$\times$) using a sandwich training strategy that samples minimum, maximum, and random widths.
\textit{MatFormer}~\cite{matformer} extended this concept to transformers with Matryoshka-style nested MLP widths, demonstrating up to 2$\times$ speedup with minimal accuracy degradation. However, MatFormer's elasticity is limited to MLP hidden dimensions, while attention and feature extractor remain fixed. Despite their success in artificial neural networks, elastic architectures have received little attention in the context of SNN. Directly transferring elasticity concepts to SNNs is non-trivial, as architectural scaling in SNNs affects not only parameter counts but also spike dynamics and synaptic activity. To the best of our knowledge, no prior work has explored a fully elastic spiking transformer architecture designed for neuromorphic-native execution.

\

\section{ARCHITECTURE}

We propose NESTformer, an elastic spiking transformer that enables runtime granularity control via three elastic building blocks - MLP, self-attention, and feature extractor. Network design is built on three principles: (1) \textbf{Nested Architecture}, similar to MatFormer \cite{matformer}, where all smaller configurations are strict subsets of larger ones, enabling weight sharing and efficient training; (2) \textbf{Spike-Aware Elasticity} where size reduction is designed to minimize both parameters and, more importantly, spike activity; and (3) \textbf{Hardware-Aware Design} where only structures typically supported by native neuromorphic hardware are supported, avoiding CPU-based ops like BMM, leading to the design of a novel, neuromorphic-friendly attention block design.

The NESTformer architecture processes event-based input through an elastic feature extractor, followed by spiking attention-MLP elastic blocks. The granularity parameter $g$ controls the effective capacity across all three elastic components (Table \ref{tab:elastic_g}), enabling selection of a nested subnet within the entire network without the need to train separate capacity networks. The following sections detail the three different architectural building blocks.

\begin{table}[h]
\caption{Effects of granularity $g$ on network scaling}
\label{tab:elastic_g}
\centering
\small
\begin{tabular}{lcccc}
\toprule
\textbf{Component} & \textbf{g0} & \textbf{g1} & \textbf{g2} & \textbf{g3} \\
\midrule
MLP Hidden Width & 64 & 160 & 416 & 1024 \\
Attention Heads & 4 & 8 & 16 & 32 \\
Conv Channels & 16 & 32 & 64 & 128 \\
Total Parameters & 0.68M & 0.92M & 1.46M & 2.59M \\
\bottomrule
\end{tabular}
\end{table}

\subsection{Elastic Spiking Self-Attention}
We propose an elastic spiking self-attention mechanism based on dynamic head slicing, inspired by Matryoshka representation learning \cite{matryoshka_style_representation_learning}. During inference, we dynamically slice the Query ($Q$), Key ($K$), and Value ($V$) tensors to the active head count $h_g$:
\begin{equation}
Q_g, K_g, V_g = {Q, K, V}[:, :, :h_g, :, :]
\end{equation}

Additionally, we modify the attention mechanism to ensure compatibility with neuromorphic hardware. Standard spiking transformers often compute attention maps using standard general matrix multiplications (GEMM) operations. NESTformer introduces an explicit intermediate LIF activation between the $Q \cdot K^T$ and the $V$ projection. This converts the attention scores into binary spikes, ensuring that the subsequent multiplication with $V$ is a strictly spike-driven accumulation:

\begin{equation}
\colorbox{acolor}{$\text{attn}_g$} = \text{LIF}(\colorbox{qcolor}{$Q_g$} \cdot K_g^T), \quad \colorbox{ocolor}{$\mathbf{O}_g$} = \text{LIF}(\text{attn}_g \cdot V_g)
\end{equation}

This allows mapping the attention mechanism natively to event-driven hardware described in the following section.

\begin{figure}[htbp]
  \centering
  \includegraphics[width=\linewidth]{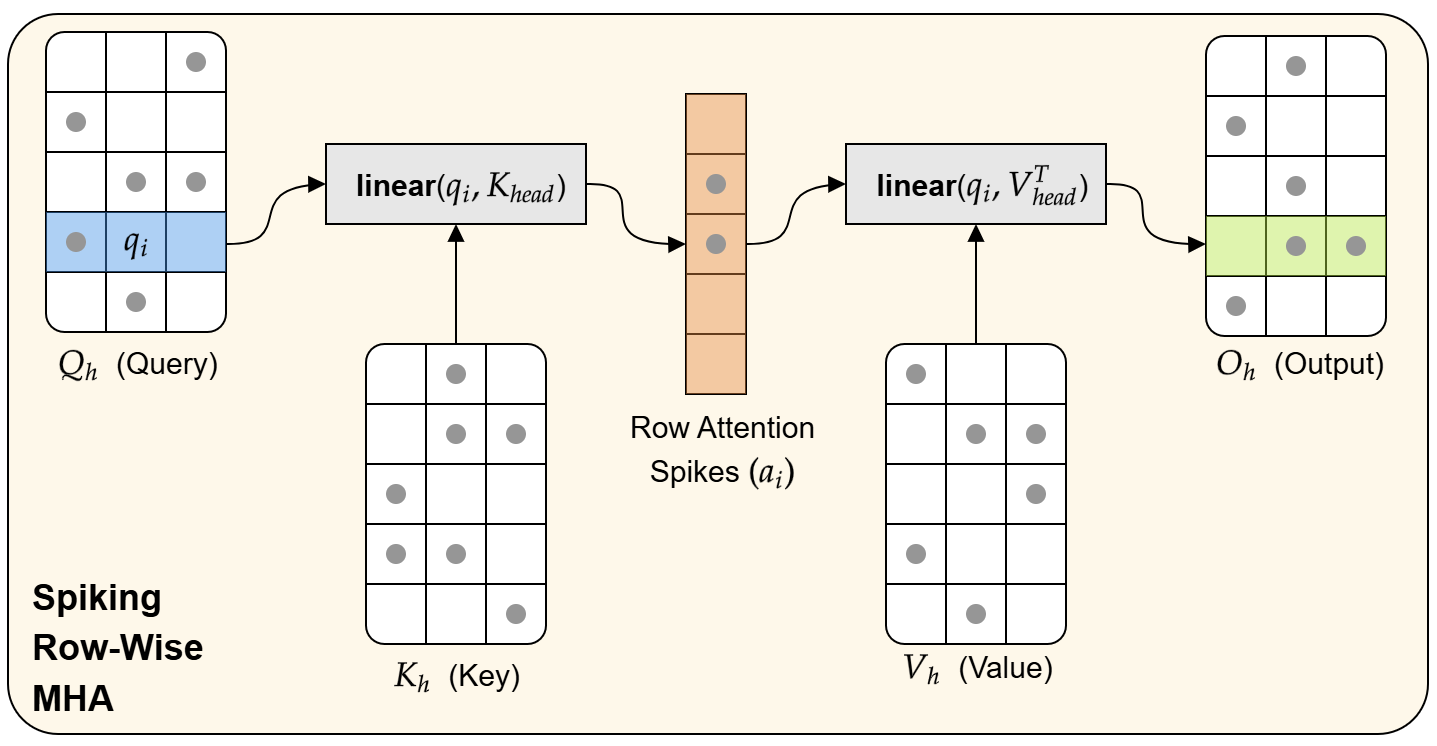}
  \caption{Proposed spiking row-wise attention module}
  \label{fig:sr_mha}

  \vspace{1em} 
  \hrule       
  \vspace{0.5em}

  \captionof{algorithm}{Row-wise Attention Computation}
  \label{alg:row_wise_attn}
\hrule
\vspace{0.5em}
  \begin{algorithmic}[1]
    \small 
    \State $O \gets \text{zeros}(T, B, H, N, D)$
    \For{$t, b, h$ \textbf{in} $T, B, H$}
        \State $K_{head}, V_{head} \gets K[t, b, h], V[t, b, h]$
        \For{$i = 1$ \textbf{to} $N$}
            \State \colorbox{qcolor}{$q_i \gets Q[t, b, h, i]$}
            \State \colorbox{acolor}{$a_i \gets \text{LIF} \circ \text{linear}(q_i, K_{head})$}
            \State \colorbox{ocolor}{$o_i \gets \text{LIF} \circ \text{linear}(a_i, V_{head}^\top)$}
            \State $O[t, b, h, i] \gets o_i$
        \EndFor
    \EndFor
    \State \Return $O$
  \end{algorithmic}
  \hrule 
\end{figure}

\subsection{Spiking Row-wise Attention}

Standard spiking attention mechanisms rely on GEMM \cite{cml} or masking operations \cite{qkformer, spikformer} to compute attention scores. While highly optimized for GPUs, these operations require generating and storing dense intermediate matrices, making them fundamentally incompatible with spike-driven neuromorphic hardware (e.g., Loihi \cite{intel_loihi}, SpiNNaker~\cite{spinnaker}, Spiker~\cite{10794606}). True neuromorphic architectures operate only by accumulating spikes locally in spike-driven LIF units and lack a global memory to process and store these intermediate dense matrices. Consequently, existing implementations often offload this `middle step' to a coprocessor or CPU, which processes these matrices densely at each timestep, negating the energy benefits of the accelerator. To solve this, NESTformer introduces a row-wise formulation that decomposes the monolithic matrix multiplication into a sequence of two LIF operations. The standard attention computation can be easily split row-by-row through a simple mathematical transformation:

\begin{equation}
     O_{h} = Q_h (K_h^T V_{h}) = \begin{bmatrix}
         (q_{h1} K_h^T) V_h \\
         (q_{h2} K_h^T) V_h \\
         \vdots\\
         (q_{hd} K_h^T) V_h \\
     \end{bmatrix}  \text{ for each } h
  \label{eq:important}
\end{equation}

For each row in $O_h$, this operation is equivalent to a sequence of two linear LIF layers, using $K_h^T$ and $V_h$ as weights. This allows deployment on native neuromorphic hardware without requiring offloading to a support CPU. This changes the way the attention operation is mapped in hardware into a very standard operator, supported by the majority of currently available neuromorphic chips.
Algorithm~\ref{alg:row_wise_attn} shows the code for this simple transformation. It should be noted that this formulation is mathematically equivalent to the standard Spiking Self Attention \cite{spikformer} with the addition of the intermediate LIF operation described earlier (Algorithm \ref{alg:row_wise_attn}, line 6).
This row-wise formulation additionally makes implementing elasticity trivial, as adjusting the granularity $h_g$ simply changes the loop limit of the outer iterator (Line 2).
While row-wise processing incurs a 2.6--10.2$\times$ slowdown on GPUs (due to loss of BLAS parallelization), our formulation is mathematically equivalent to the parallel GEMM form. This allows us to train efficiently on GPUs using the standard parallel form, then convert the model to the row-wise formulation for deployment.
Figure \ref{fig:sr_mha} shows a graphical representation of the proposed row-wise attention computation.

\subsection{Elastic MLP}

Similar to Matformer \cite{matformer}, the MLP module is designed to dynamically adjust hidden dimension using log-spaced filter counts. In standard transformers, MLP layers account for approximately 60\% of parameters. As a result, an elastic MLP can significantly reduce the network parameter count, enabling deployment on more resource-constrained neuromorphic hardware without retraining. The granularities follow a log-spacing to provide roughly uniform capacity increments:
\begin{equation}
h_g = \text{round}\left(h_{\min} \cdot 2^{g \cdot \log_2(h_{\max}/h_{\min})/(G-1)}\right)
\end{equation}

where $h_g \in \{64, 160, 416, 1024\}$ is the MLP hidden dimension for granularity $g$.

\begin{figure}[]
\centering
\includegraphics[width=\linewidth]{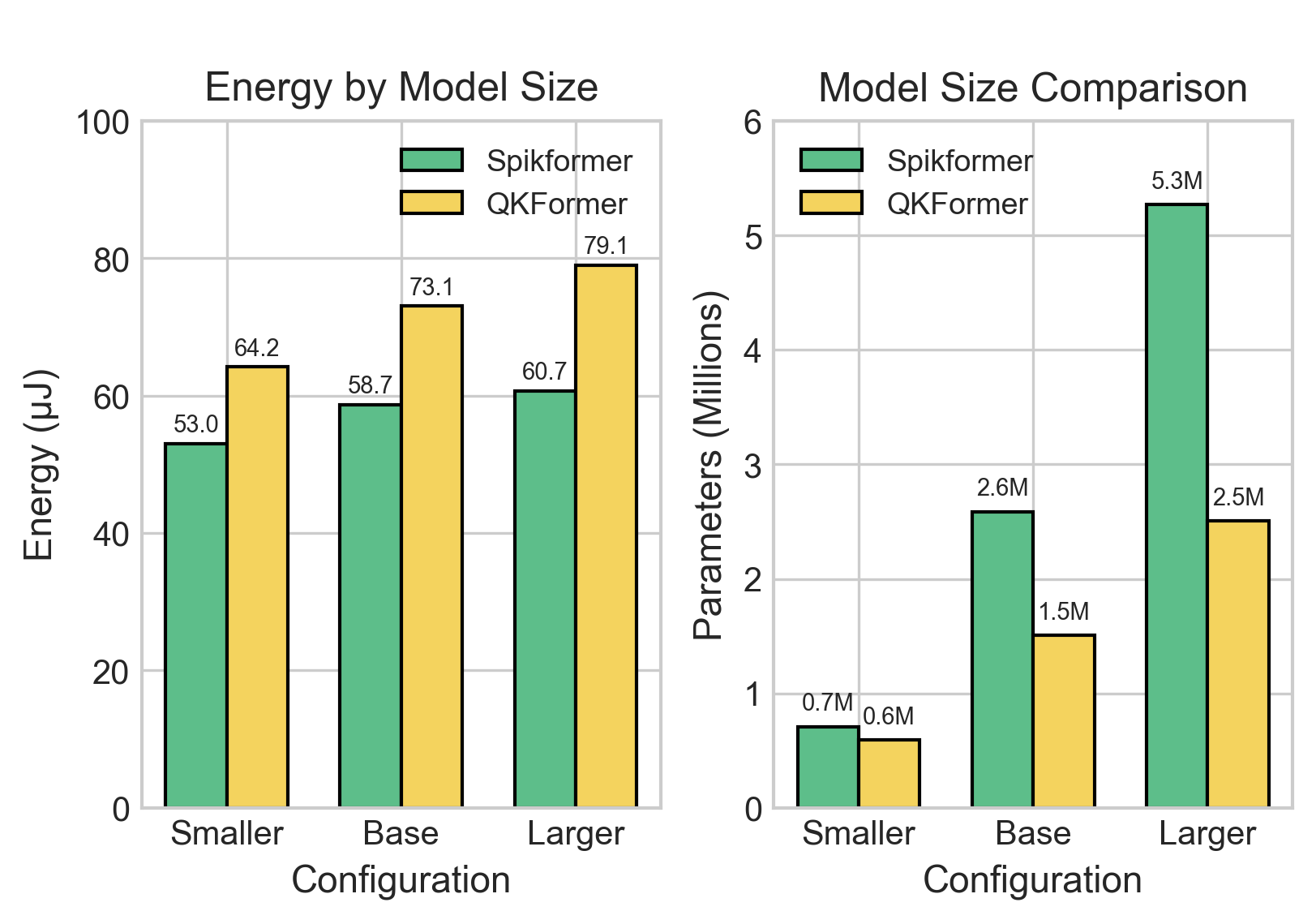}
\caption{Scaling analysis of sota spiking transformers. While parameter counts scale significantly between ``Larger" and ``Smaller" configurations (Right), the corresponding energy consumption remains high (Left).}
\label{fig:energy_param_scaling}
\end{figure}

\subsection{Elastic Patch Embedding}

To motivate the design of our feature extractor, we first analyze the scalability of current state-of-the-art spiking transformers, specifically Spikformer~\cite{spikformer} and QKFormer~\cite{qkformer}. By varying the block count, embedding dimension, and number of attention heads, we generated ``S", ``B", and ``L" variants of these baselines. We measure total spikes per inference, and we estimate energy consumption using Intel Loihi's energy model (23.6 pJ per synaptic operation). As illustrated in Figure~\ref{fig:energy_param_scaling}, these networks show a distinct disconnect between parameter scaling and energy efficiency. While reducing model dimensions successfully scales the memory footprint, the corresponding energy savings are disproportionately low, dropping by only $\sim$12\% (60.7 $\mu$J to 53.0 $\mu$J) when parameter count varies by over 80\% (5.3M to 0.7M).

\begin{figure}[h]
\centering
\includegraphics[width=\linewidth]{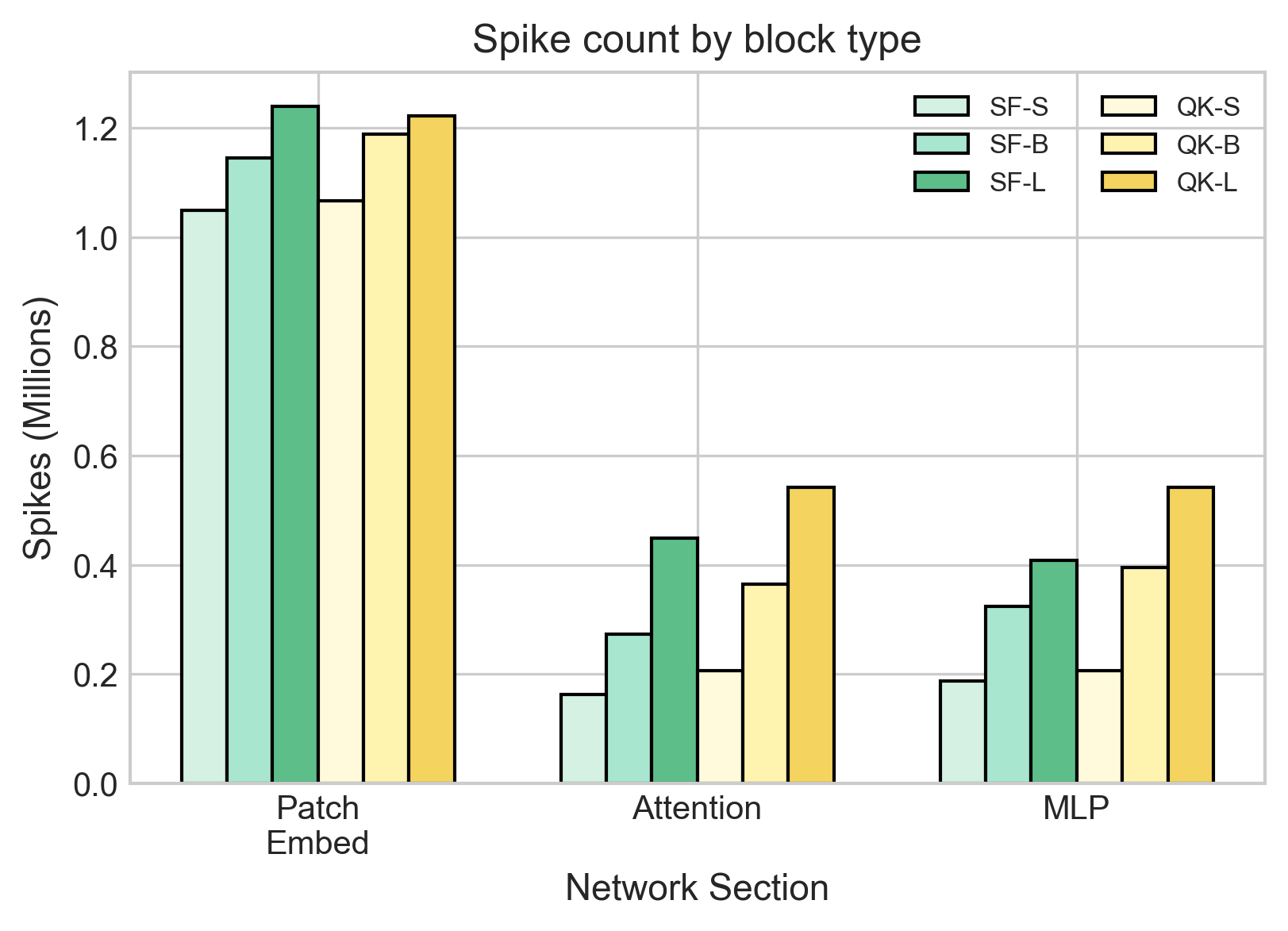}
\includegraphics[width=\linewidth]{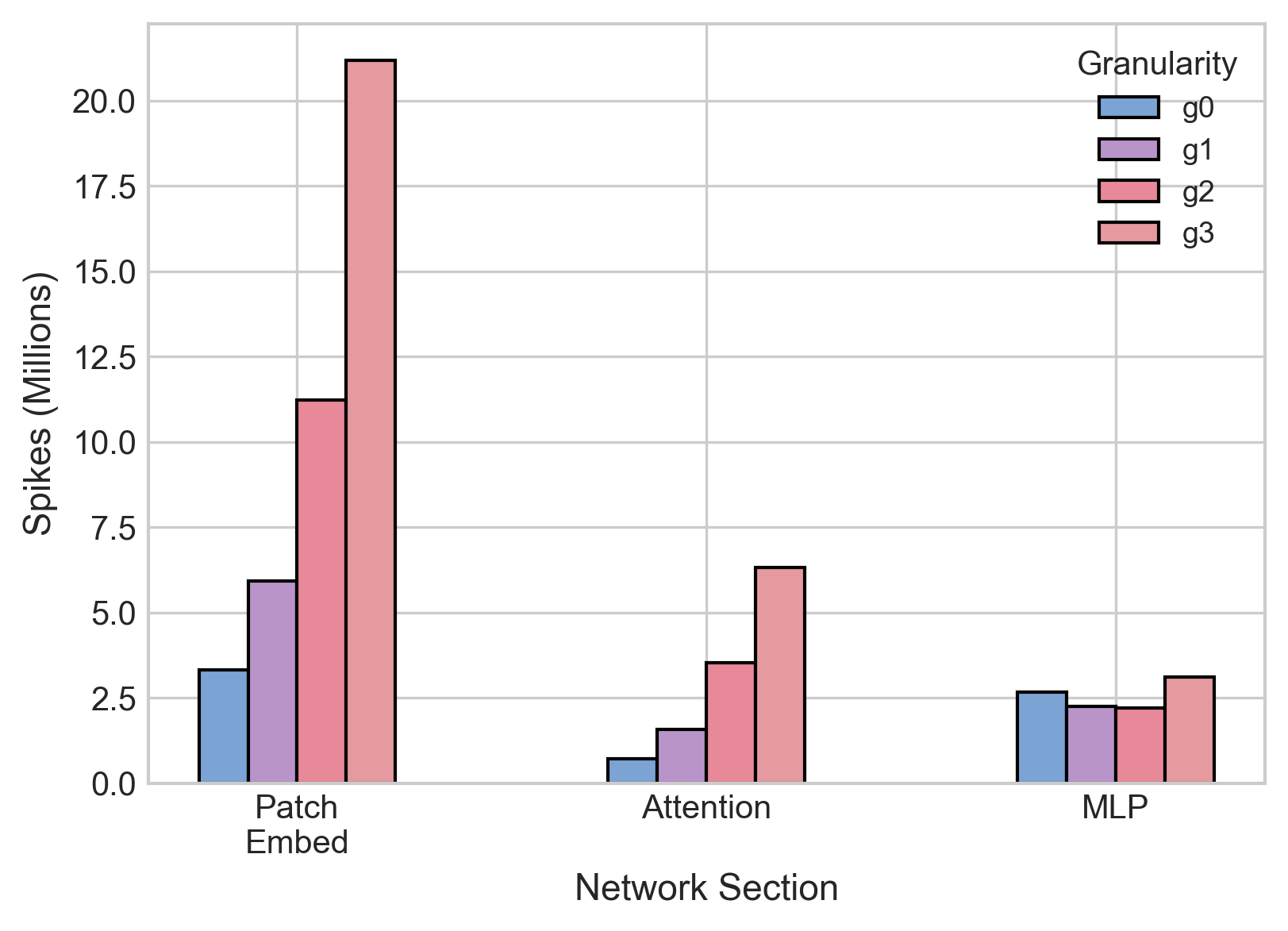}
\caption{Spike count distribution by network section. In standard spiking transformer architectures (top), the patch embedding stage (left bars) accounts for the majority of activity, generating significantly more spikes than the attention or MLP blocks combined. Notably, this stage does not scale down proportionally with the rest of the model architecture. NESTformer (bottom) shows instead much better spike scaling at different granularities}
\label{fig:spike_distribution}
\end{figure}

\begin{figure}[]
  \centering
   \includegraphics[width=\linewidth]{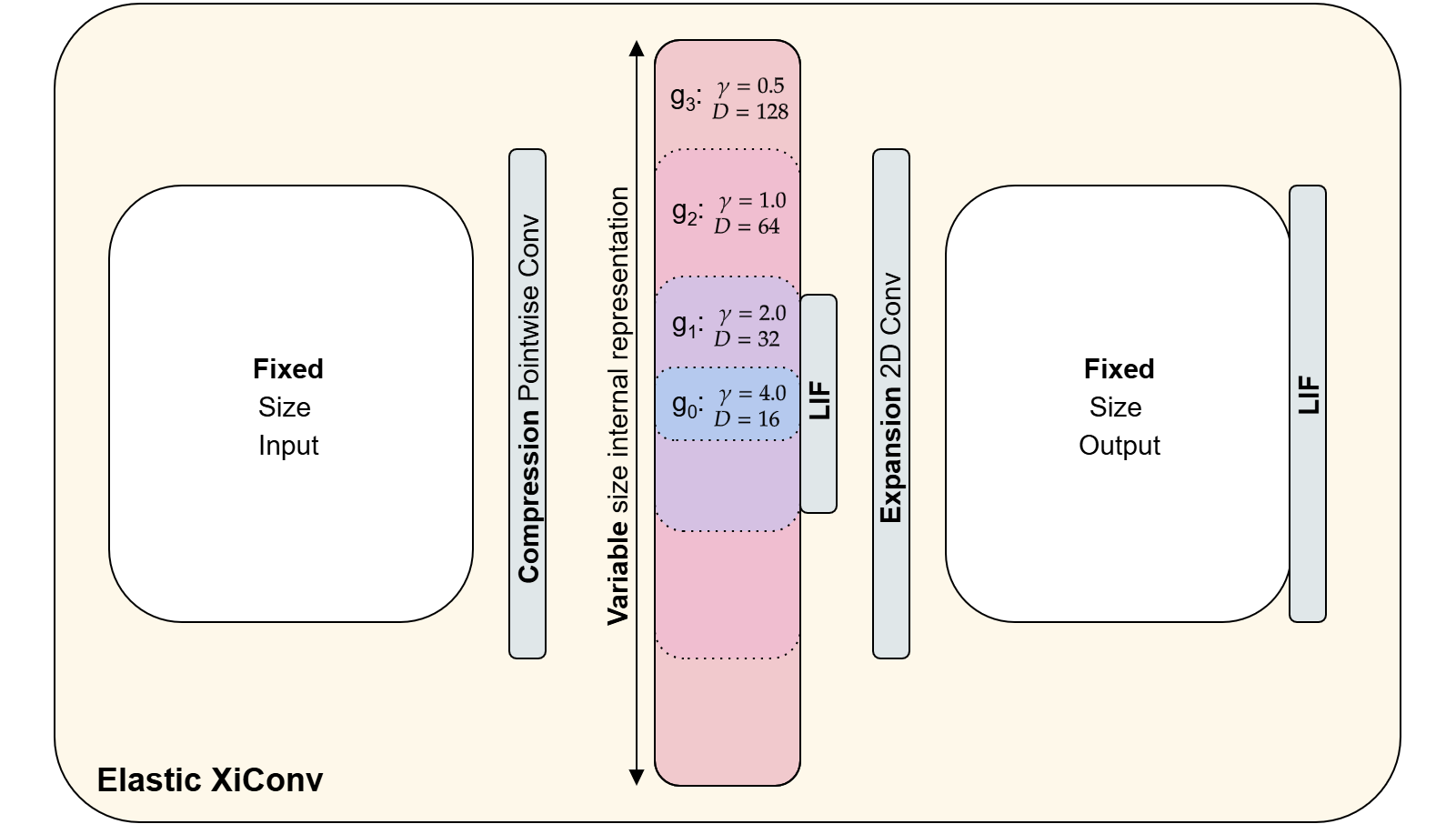}
   \caption{Proposed patch embedding elastic block}
   \label{fig:xissa}
\end{figure}

Figure~\ref{fig:spike_distribution} (top) breaks down the total spike generation by network section. The analysis shows that the patch-embedding stage is the primary bottleneck for spike generation. Because this stage processes feature maps at the highest spatial resolution, it accounts for the vast majority of network activity, consistently generating over $1.0\times 10^6$ spikes across model sizes. In contrast, the downstream attention and MLP blocks operate on tokenized, lower-dimensional representations and contribute significantly less to the total spike count. Consequently, standard scaling strategies that reduce the transformer depth or width fail to address the dominant energy cost incurred at the input stage. To address this specific bottleneck, we introduce a novel elastic spiking patch splitting (XiSPS) backbone. Unlike standard static embeddings, XiSPS supports dynamic, runtime trade-offs between feature granularity and power consumption without retraining. We adapt the scalability mechanism from the XiConv block \cite{xinet} as our foundation, leveraging a tunable compression factor $\gamma$ to reduce both computational load and parameter count. The novel XiSPS block is designed using LIF neurons to ensure purely event-driven processing. To simplify training and the elasticity mechanism, we maintain constant input and output filter counts while dynamically modifying the compressed feature map size during training. The final block architecture used in the patch embedding stage is shown in Figure \ref{fig:xissa}. Figure \ref{fig:nest_granularity_p_acc_e} shows the effect that the optimized backbone scaling has on energy usage, demonstrating a stronger correlation between network size and inference energy.

\begin{table}[]
\centering
\caption{Spike activity and energy scaling across g}
\label{tab:spike_activity}
\begin{tabular}{lcccc}
\toprule
  & \textbf{Spikes/Inf} & \textbf{Firing Rate} & \textbf{Energy ($\mu$J)} & \textbf{Relative} \\
\midrule
g0 & 0.56M & 1.04\% & 13.3 & 29\% \\
g1 & 0.82M & 1.49\% & 19.3 & 42\% \\
g2 & 1.22M & 2.19\% & 28.7 & 62\% \\
g3 & 1.95M & 3.39\% & 46.1 & 100\% \\
\bottomrule
\end{tabular}
\end{table}

\subsection{Granularity-Energy Relationship and Spike Dynamics}
To validate the efficacy of our elastic design, we analyze the relationship between model granularity, total spike activity, and energy consumption. As shown in Table \ref{tab:spike_activity}, the results confirm that reducing granularity directly leads to energy savings, unlike previous architectures. Moving from the largest configuration ($g3$) to the smallest ($g0$) results in a 71\% decrease in total spike events (1.95M $\rightarrow$ 0.56M) and a proportional reduction in total energy (46.1 $\mu$J $\rightarrow$ 13.3 $\mu$J). However, a deeper layer-wise analysis reveals a non-linear dynamic within the Elastic MLP blocks.

\begin{table}[h]
\centering
\caption{Firing Rates (\%) in MLP layers across g}
\label{tab:inverse_fr}
\begin{tabular}{lcccc}
\toprule
\textbf{Layer} & \textbf{g0} & \textbf{g1} & \textbf{g2} & \textbf{g3} \\
\midrule
block.0.mlp.lif1 & 7.46\% & 6.67\% & 5.47\% & 2.69\% \\
block.0.mlp.lif2 & 6.67\% & 7.56\% & 6.05\% & 2.96\% \\
block.1.mlp.lif1 & 20.96\% & 13.66\% & 8.00\% & 4.14\% \\
block.1.mlp.lif2 & 35.52\% & 25.71\% & 12.78\% & 11.31\% \\
\bottomrule
\end{tabular}
\end{table}

We observe an inverse relationship between the MLP hidden dimension and average firing rate, which can be seen in Figure \ref{fig:spike_distribution}. As illustrated in Table \ref{tab:inverse_fr}, as the granularity reduces the hidden dimension from 1024 ($g3$) to 64 ($g0$), the average firing rate of the remaining neurons increases significantly (from $\sim$4.38\% to $\sim$21.01\%).

\begin{table*}[t]
\centering
\caption{Comparison of Energy, Complexity, and Accuracy on EHWGesture. T=16 for all approaches.}
\label{tab:ehwgesture_results}
\setlength{\tabcolsep}{8pt} 
\begin{tabular}{l c c c c c}
\toprule
\textbf{Model} & \textbf{Energy ($\mu J$)} $\downarrow$ & \textbf{Param (M)} $\downarrow$ & \textbf{Accuracy (\%)} $\uparrow$ & \textbf{AQA (\%)} $\uparrow$ & \textbf{Gesture (\%)} $\uparrow$ \\
\midrule
\multicolumn{6}{c}{\textit{CNN Baselines}} \\
\midrule
ResNet50 & 147,000 & 25.60 & 65.52 & 71.48 & 91.02 \\
ResNeXt & 460,000 & 116.10 & 68.51 & 72.58 & 94.41 \\
PhiNet & 6,900 & 4.91 & 71.25 & 75.11 & \textbf{96.98} \\
\midrule
\multicolumn{6}{c}{\textit{SNN Baselines}} \\
\midrule
Spikformer-1-160 & 55.3 & 0.71 & 71.89 & 75.94 & 96.20 \\
Spikformer-2-256 & 59.0 & 2.60 & 73.62 & 78.29 & 95.86 \\
Spikformer-3-320 & 70.8 & 5.27 & 74.16 & 79.00 & 96.40 \\
QKFormer-2-160 & 48.1 & 0.69 & 73.58 & 77.59 & 96.32 \\
QKFormer-4-256 & 73.6 & 1.51 & 75.11 & 78.87 & 96.22 \\
QKFormer-5-320 & 100.9 & 2.35 & 75.28 & 78.92 & 96.86 \\
\midrule
\multicolumn{6}{c}{\textit{Ours}} \\
\midrule
\textbf{NESTformer (g3)} & 46.1 & 2.59 & \textbf{75.98} & \textbf{79.45} & 96.40 \\
\textbf{NESTformer (g2)} & 28.7 & 1.46 & 75.86 & 78.92 & 96.32 \\
\textbf{NESTformer (g1)} & 19.3 & 0.92 & 72.38 & 74.66 & 95.66 \\
\textbf{NESTformer (g0)} & \textbf{13.3} & \textbf{0.68} & 71.01 & 74.04 & 94.78 \\
\bottomrule
\end{tabular}
\end{table*}

\subsection{Training}
Unlike standard slimmable approaches that utilize sandwich training (aggregating losses from minimum, maximum, and random sub-networks), we adopt a simpler stochastic sampling strategy that updates a single granularity configuration per training step. We empirically found that uniform sampling often leaves larger-granularity networks under-trained. To counteract this, we sample the active granularity $g$ from a biased distribution $P(g)$ shifted toward larger configurations: $g_t \sim P(g), $ where P depends on the number of parameters per granularity setting. This ensures that the full parameter set receives frequent updates, stabilizing the shared weights that support all smaller subnetworks. To maintain high throughput during training and avoid the significant overhead of dynamic graph recompilation, we rely on static tensor shapes. We allocate all elastic tensors to their maximum capacity $g_{\max}$ and apply binary masks to zero-pad inactive regions for lower sampled granularities. This guarantees a consistent computational graph structure and memory layout throughout training. This masking strategy, combined with per-granularity batch normalization, enables different configurations to be trained efficiently within a single model. Training uses the AdamW optimizer with a learning rate of $1\mathrm{e}{-3}$, cosine annealing, and weight decay of $6\mathrm{e}{-2}$. Compared to training a single-size network, we empirically observed that increasing the number of training steps by around 50\% was required to achieve convergence across all four sub-networks.

\section{EXPERIMENTS}

We validate NESTformer on a variety of event-driven and static benchmarks, with a primary focus on gesture understanding for clinical and human-computer interaction (HCI) applications. Our primary evaluation uses the new EHWGesture dataset, where NESTformer outperforms the proposed baselines while using significantly less energy. In addition to EHWGesture, we report results on standard neuromorphic benchmarks (DVS128 Gesture, CIFAR10-DVS) and static classification tasks (CIFAR-10/100) to demonstrate the generalization capabilities for the proposed architecture.

\begin{figure*}[]
  \centering
   \includegraphics[width=.9\linewidth]{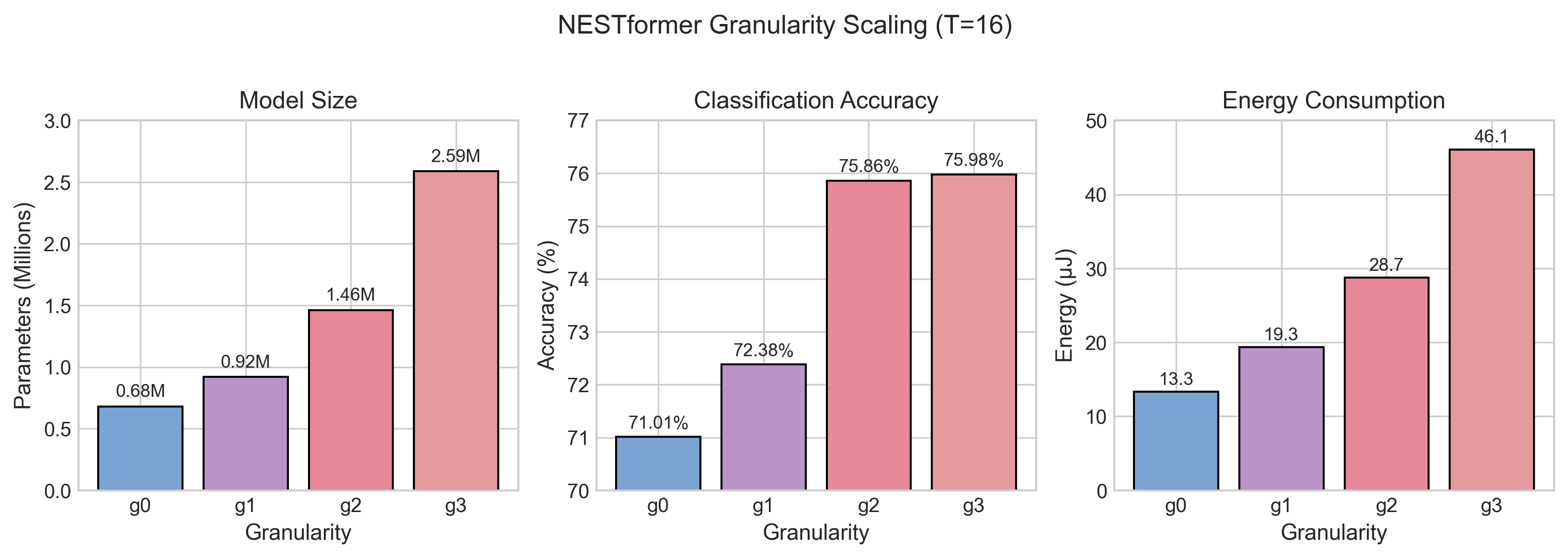}
   \caption{Effects of different granularity levels on network parameter count, accuracy, and energy consumption}
   \label{fig:nest_granularity_p_acc_e}
\end{figure*}

\subsection{EHWGesture: Clinical Gesture Recognition}
\begin{figure}[h]
  \centering
   \includegraphics[width=\linewidth]{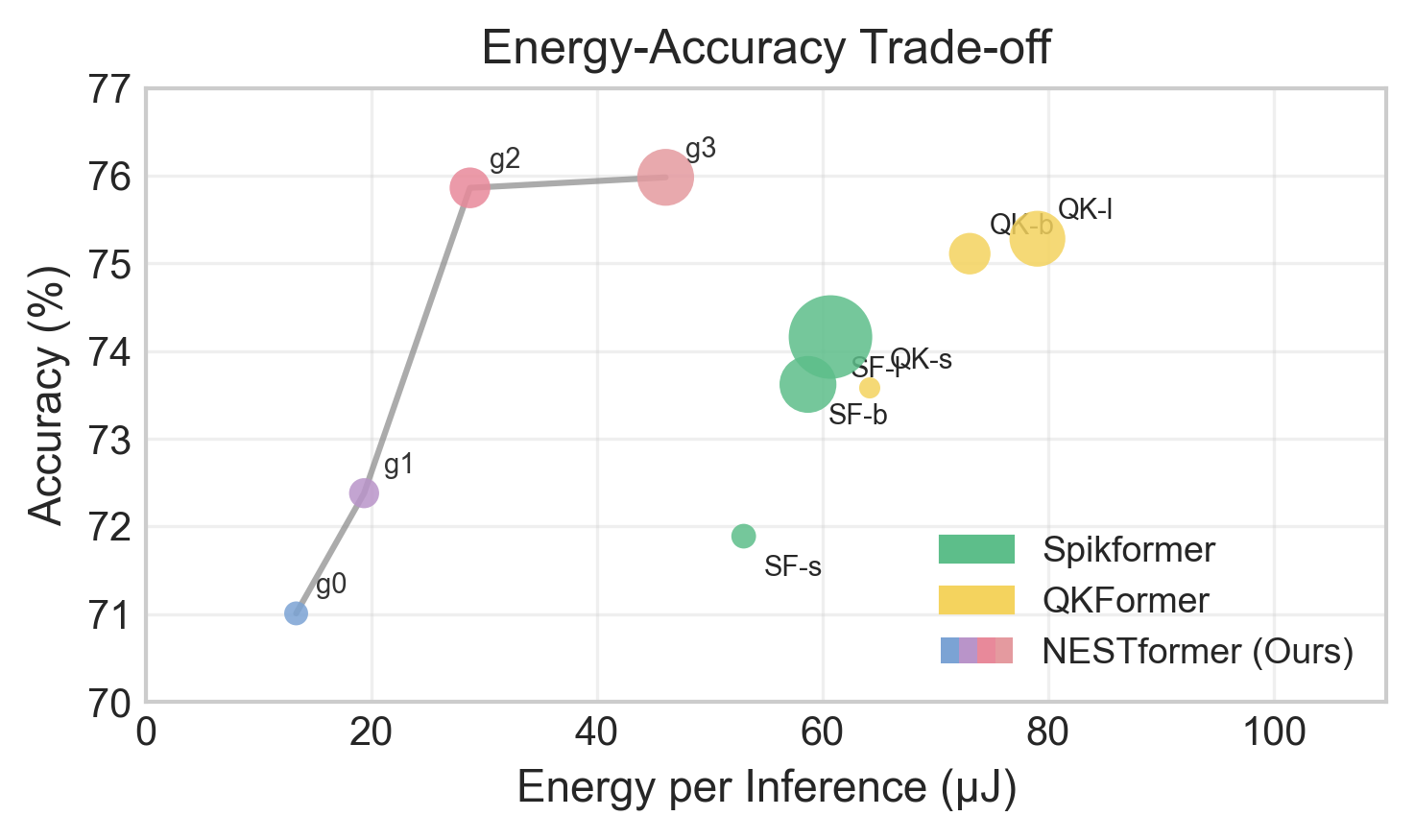}
   \caption{Energy / Accuracy tradeoff for the proposed approach compared to alternative SNN baselines}
   \label{fig:nest_acc_vs_energy}
\end{figure}

The EHWGesture dataset serves as our primary evaluation benchmark, comprising 9,708 samples across 11 clinical gesture classes. As detailed in Table~\ref{tab:ehwgesture_results}, NESTformer establishes a new state-of-the-art for event-based gesture recognition, surpassing both static SNNs and efficient ANN baselines. Our largest configuration (g3) achieves a peak accuracy of 75.98\%, outperforming the strongest baseline, QKFormer-5-320 (75.28\%), and the lightweight PhiNet CNN (71.25\%). This performance advantage is coupled with significant efficiency gains; NESTformer (g3) consumes only 46.1 $\mu$J per inference, less than half the energy of QKFormer-5-320 (100.9 $\mu$J) and orders of magnitude less than the CNN baselines, which range from 6,900 $\mu$J to over 460,000 $\mu$J. The architecture also demonstrates good flexibility through its elastic granularities. The intermediate g2 configuration emerges as a sweet spot for general deployment, retaining 99.8\% of the peak accuracy (75.86\%) while reducing energy consumption by nearly 40\% to 28.7 $\mu$J. For scenarios demanding extreme energy conservation, such as always-on monitoring, the g1 and g0 configurations provide an ultra-low-power alternative. They maintain a competitive accuracy comparable to the best baseline in the original dataset paper, while consuming 19.3 to 13.3 $\mu$J, enabling continuous monitoring on battery-operated devices. Figure \ref{fig:nest_acc_vs_energy} shows the energy-accuracy tradeoffs that can be reached with the proposed architecture. In contrast, Figure \ref{fig:nest_granularity_p_acc_e} details the effects of granularity scaling on model size, accuracy, and energy.

\begin{table*}[]
\centering
\caption{Comparison with State-of-the-Art across Event-Based and Static Benchmarks. ``-'' indicates the metric was not reported or applicable. Best results are \textbf{bolded}.}
\label{tab:overall_benchmarks}
\footnotesize
\setlength{\tabcolsep}{4pt}
\begin{tabular}{l | c c c | c c c | c c c c}
\toprule

\multirow{3}{*}{\textbf{Model}} & \multicolumn{3}{c|}{\textbf{DVS128 Gesture}} & \multicolumn{3}{c|}{\textbf{CIFAR10-DVS}} & \multicolumn{4}{c}{\textbf{Static (CIFAR-10 / 100)}} \\
 & \textbf{Params} $\downarrow$ & \textbf{Timestep} & \textbf{Accuracy} $\uparrow$ & \textbf{Params} $\downarrow$ & \textbf{Timestep} & \textbf{Accuracy} $\uparrow$ & \textbf{Params} $\downarrow$ & \textbf{Timestep} & \textbf{C10 Acc} $\uparrow$ & \textbf{C100 Acc} $\uparrow$ \\
 & \textbf{(M)} & \textbf{(T)} & \textbf{(\%)} & \textbf{(M)} & \textbf{(T)} & \textbf{(\%)} & \textbf{(M)} & \textbf{(T)} & \textbf{(\%)} & \textbf{(\%)} \\

\midrule
\multicolumn{11}{c}{\textit{SNN Baselines}} \\
\midrule

Spikformer \cite{spikformer} & 2.59 & 16 & 98.96 & 2.57 & 16 & 80.90 & 9.32 & 4 & 95.51 & 78.21 \\

QKFormer \cite{qkformer} & 1.50 & 16 & 98.60 & 1.50 & 16 & \textbf{84.00} & 6.74 & 4 & \textbf{96.18} & 81.15 \\

Spike-driven TF \cite{spike-driven-transformer} & 1.05 & 16 & \textbf{99.30} & 1.05 & 10 & 77.40 & 9.32 & 4 & 95.51 & 77.92 \\

STSA \cite{stsa} & 1.99 & 16 & 98.70 & 1.99 & 16 & 79.93 & - & - & - & - \\

Spikingformer \cite{cml} & 2.57 & 16 & 98.60 & 2.57 & 16 & 80.90 & 9.32 & 4 & 96.04 & 80.02 \\

SEW-ResNet \cite{sew_resnet} & - & 16 & 97.90 & - & - & - & - & 4 & 94.85 & 74.24 \\

LIAF-Net \cite{liaf_net} & - & 60 & 97.60 & - & 10 & 70.40 & - & - & - & - \\

PLIF \cite{plif} & 17.40 & 20 & 97.60 & 17.40 & 20 & 74.80 & - & - & - & - \\

TET (ResNet-19) \cite{tet} & - & - & - & 12.63 & 4 & 74.47 & 12.63 & 4 & 94.44 & 74.47 \\

Dspike \cite{dspike} & - & 10 & 96.90 & - & 10 & 75.40 & - & - & - & - \\

\midrule
\multicolumn{11}{c}{\textit{NESTformer (Ours)}} \\
\midrule

\textbf{NESTformer (g3)} & 2.07 & 16 & \textbf{99.30} & 1.87 & 16 & 83.61 & 4.14 & 4 & \textbf{96.18} & \textbf{81.30} \\

\textbf{NESTformer (g2)} & 1.21 & 16 & 98.61 & 0.98 & 16 & 82.52 & 2.36 & 4 & 95.91 & 80.12 \\

\textbf{NESTformer (g1)} & 0.63 & 16 & 98.26 & 0.56 & 16 & 80.94 & 1.88 & 4 & 93.25 & 79.24 \\

\textbf{NESTformer (g0)} & \textbf{0.39} & 16 & 97.57 & \textbf{0.28} & 16 & 79.88 & \textbf{0.98} & 4 & 92.83 & 76.11 \\

\bottomrule
\end{tabular}
\end{table*}

\subsection{Architecture generalization}

In addition to EHWGesture, we evaluate NESTformer on two standard event-based benchmarks (DVS128 Gesture~\cite{dvs128_gesture}, CIFAR10-DVS~\cite{cifar10_dvs}) and two static image classification benchmarks (CIFAR-10~\cite{cifar10}, CIFAR-100~\cite{cifar100}). The results, summarized in Table~\ref{tab:overall_benchmarks}, demonstrate that our elastic framework matches or outperforms fixed-topology state-of-the-art models while offering significant energy savings.

On DVS128 Gesture, NESTformer (g3) matches the peak accuracy of Spike-Driven Transformer~\cite{spike-driven-transformer} (99.30\%) and outperforms QKFormer (98.60\%). Notably, the ultra-lightweight g0 configuration retains 97.57\% accuracy, higher than several specialized baselines, while utilizing 5.3$\times$ fewer parameters than the g3 configuration. On the more challenging CIFAR10-DVS classification task, NESTformer (g3) achieves 83.61\%, remaining highly competitive with QKFormer (84.00\%) and outperforming standard Spikformer (80.90\%). The benefits of elasticity are most evident here: the g2 configuration (0.98M params) achieves 82.52\%, delivering 99\% of the SOTA performance with 35\% fewer parameters than QKFormer. Finally, on static CIFAR-10/100 datasets, NESTformer demonstrates that neuromorphic elasticity generalizes to frame-based data. Notably, on CIFAR-100, our g3 configuration (81.30\%) achieves a new state-of-the-art among spiking transformers, surpassing both QKFormer (81.15\%) and Spikformer (78.21\%).

\section{VARIABLE TIMESTEPS TRADEOFFS}

We analyze the tradeoffs between temporal resolution (timesteps $T$), model granularity ($g$), and task-specific performance on the EHWGesture dataset. Our analysis highlights the distinct temporal dependencies between gesture classification and action quality assessment (AQA).

\subsection{Temporal Resolution vs. Granularity}

\begin{figure*}[t]
  \centering
   \includegraphics[width=.9\linewidth]{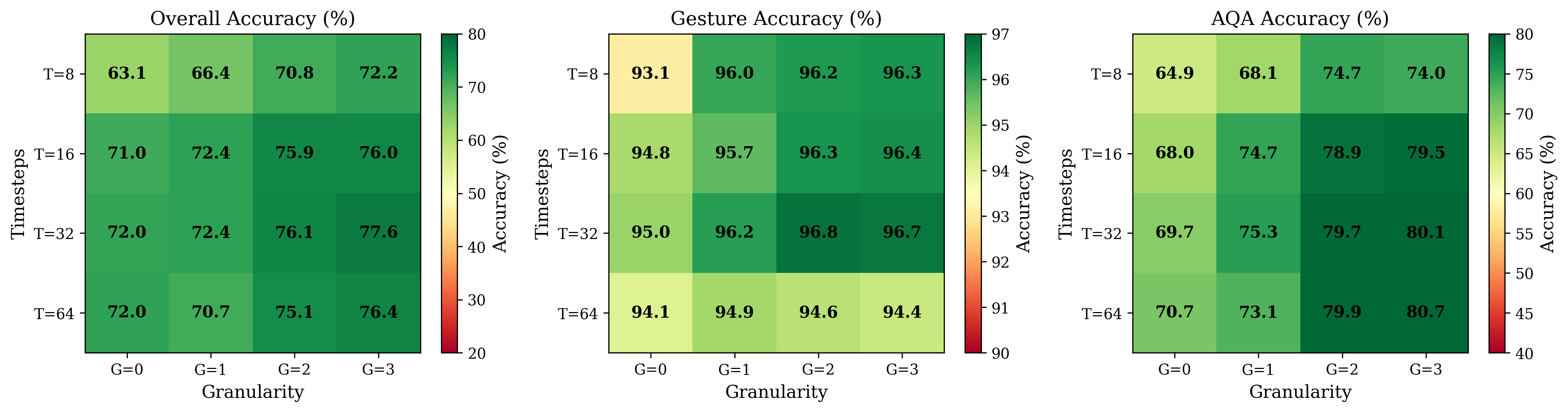}
   \caption{Accuracy heatmaps across timesteps and granularities. Left: Overall top-1 accuracy. Center: Gesture recognition accuracy. Right: Action quality assessment (AQA) accuracy. Higher granularity consistently improves performance, while optimal timesteps vary by task.}
   \label{fig:accuracy_heatmap}
\end{figure*}

Figure~\ref{fig:accuracy_heatmap} presents accuracy heatmaps across timesteps ($T \in \{8, 16, 32, 64\}$) and granularities ($g \in \{0, 1, 2, 3\}$). We find that AQA and gesture recognition exhibit opposing trends regarding temporal resolution.
\begin{itemize}
\item \textbf{AQA:} Action Quality Assessment benefits consistently from increased temporal resolution. The highest AQA performance is achieved at $T=64,\ g=3$ (80.7\%), surpassing the $T=32$ equivalent (80.1\%) and significantly outperforming $T=8$ (74.0\%). These findings align with those in the EHWGesture paper and underscore the importance of finer temporal granularity to capture motion dynamics over longer windows.
\item \textbf{Gesture Recognition:} Conversely, gesture recognition performance peaks at $16$ to $32$ timesteps and degrades as $T$ increases to 64. At $g=3$, accuracy drops from 96.7\% at $T=32$ to 94.4\% at $T=64$.  This is possibly caused by the event sensor dynamics: higher timesteps may better capture motion dynamics, but longer timesteps will include more spatial information, which is useful for determining hand and finger positions.
\end{itemize}

While both tasks generally benefit from higher $g$, it is particularly useful at lower timesteps. Shifting from $g=0$ to $g=1$ at $T=8$ significantly increases both gesture accuracy (93.1\% to 96.0\%), and AQA (64.9\% to 68.1\%). This shows that a higher $g$ enables the model to learn both spatial and motion data even when temporal resolution is constrained. For very constrained networks, this shows that even a slight increase in model parameters can yield significantly higher accuracy gains than using higher timesteps.

\subsection{Energy-Accuracy-Parameter Trade-offs}
\begin{figure}[]
\centering
\includegraphics[width=.86\linewidth]{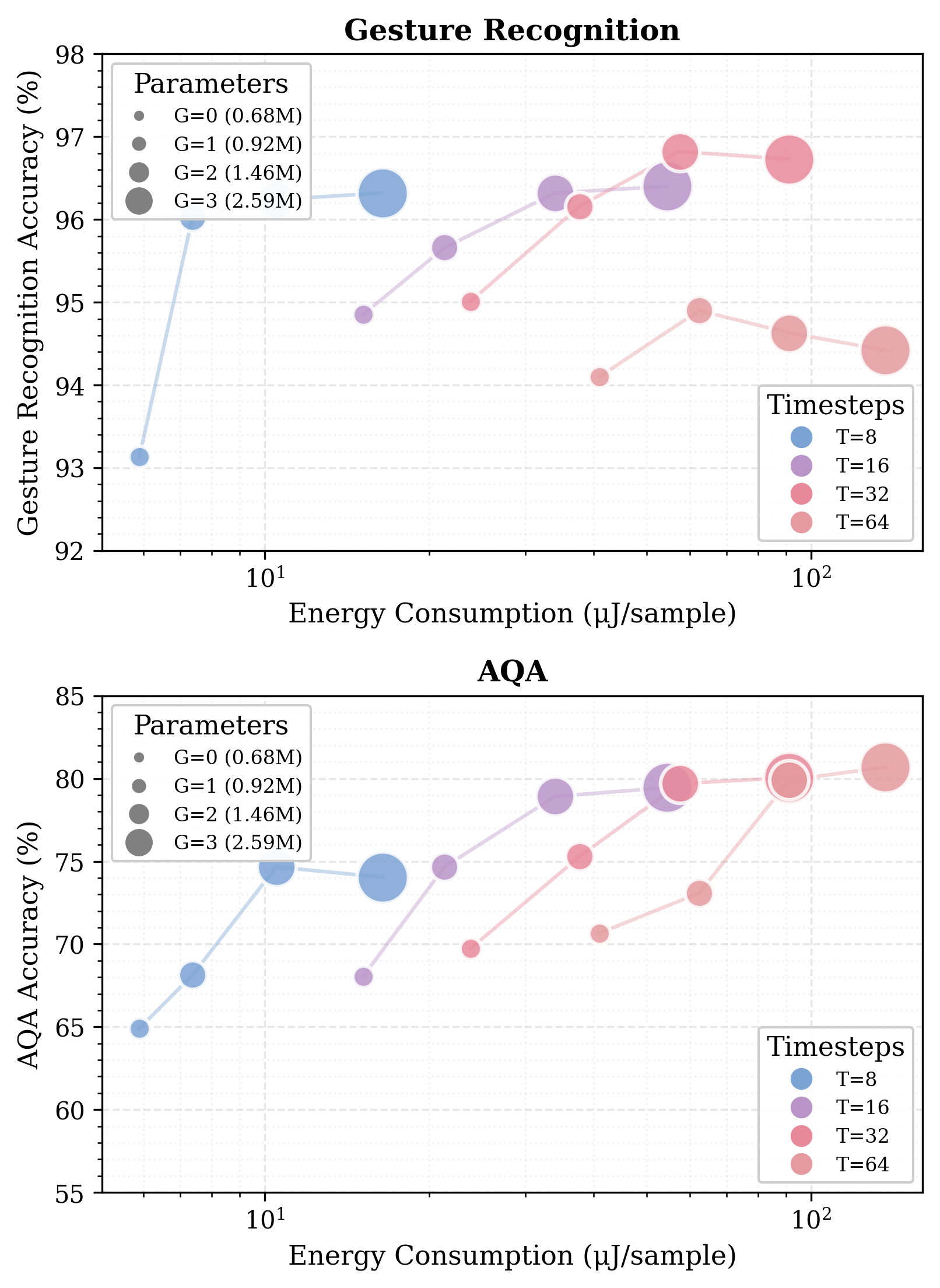}
\caption{Energy vs. accuracy trade-offs for gesture recognition (top) and AQA (bottom). Marker size indicates parameter count (0.68M--2.59M). Lines connect configurations with the same timestep. The Pareto frontier demonstrates that $T=8$ and $T=16$ configurations dominate for energy-constrained deployment.}
\label{fig:task_tradeoffs}
\end{figure}

Figure~\ref{fig:task_tradeoffs} shows the energy-accuracy trade-off. The model demonstrates distinct optimal operating points for different target tasks. For gesture recognition, the model operates most efficiently at lower timesteps. $T=8, g=2$ achieves a remarkable 96.24\% accuracy consuming only 10.49 $\mu$J. Increasing $T$ to 64 not only triples the energy to 91.17 $\mu$J (at $g=2$) but actually decreases accuracy to 94.63\%. For tasks like AQA that rely on motion dynamics, higher T offers significant advantages. The configuration $T=64, g=3$ offers the maximum AQA performance of 80.69\%. However, this comes at a steep energy cost of 136.94 $\mu$J. The $T=16, g=2$ configuration represents a balanced options achieving 96.32\% gesture accuracy and 78.92\% AQA accuracy at 28.7 $\mu$J. This configuration represents a new state-of-the-art in the accuracy-energy tradeoff for the given task.

\subsection{Spike Dynamics Across Configurations}

Table~\ref{tab:spike_dynamics} highlights the spiking behavior at granularity $g=2$. The total spike count for an inference pass scales sublinearly with both timesteps ($\sim$$2\times$ per doubling of $T$) and granularity. Notably, firing rates decrease at higher timesteps despite increased total spikes, suggesting the network distributes computation more sparsely across longer temporal windows. This can, in a real neuromorphic chip, increase overall efficiency by reducing congestion in the routing sections of the neuromorphic accelerators at each timestep, leading to higher performance than how energy is usually estimated.

\begin{table}[h]
\centering
\caption{Spike activity across configurations ($g=2$)}
\label{tab:spike_dynamics}
\small
\begin{tabular}{lcccc}
\toprule
\textbf{Timestep} & \textbf{Spikes (M)} & \textbf{Firing Rate} & \textbf{Energy ($\mu$J)} \\
\midrule
$T=8$ & 0.70 & 2.64\% & 16.5 \\
$T=16$ & 1.22 & 2.19\% & 28.7 \\
$T=32$ & 2.27 & 2.25\% & 53.5 \\
$T=64$ & 3.99 & 1.78\% & 94.2 \\
\bottomrule
\end{tabular}
\end{table}

\section{CONCLUSION}

We introduced the Elastic Spiking Transformer (NESTformer), the first spiking neural network with runtime elasticity, enabling a single trained model to operate across a 3.8$\times$ energy range without retraining. We demonstrated that our novel elastic MLP, elastic attention, and elastic patch embedding blocks enable both parameter reduction and lower energy costs with varying granularity. Our experiments demonstrate state-of-the-art accuracy on EHWGesture while providing both compatibility with neuromorphic hardware and deployment flexibility. NESTformer bridges the gap between efficient neuromorphic computing and novel elastic transformer architectures, enabling SNNs to be deployed across the full spectrum of neuromorphic hardware, from ultra-low-power (g0: 0.81 mJ) to high-accuracy edge servers (g3: 2.80 mJ) using a single unified model.

\section{DATA AVAILABILITY}

The code supporting the findings of this study is available in the public GitHub repository Elastic-Spiking-Transformers at \url{https://github.com/smilies-polito/Elastic-Spiking-Transformers}. The repository contains the project description and will host the implementation and reproducibility materials for the Elastic Spiking Transformer.


{\small
\bibliographystyle{ieee}
\bibliography{egbib}
}

\end{document}